\pdfoutput=1

\documentclass[11pt]{article}
\usepackage{ACL2023}
\usepackage{hyperref}
\usepackage{amsmath}
\usepackage{enumitem}
\usepackage{graphicx}
\usepackage{multirow}
\usepackage{algorithm}
\usepackage{algpseudocode}
\usepackage{float} 
\usepackage{microtype}
\usepackage{inconsolata}
\usepackage[utf8]{inputenc}
\usepackage[T1]{fontenc}
\usepackage{times}
\usepackage{latexsym}
\title{Prompting a Weighting Mechanism into LLM-as-a-Judge in Two-Step: A Case Study}

\author{Wenwen Xie\textsuperscript{1}, Gray Gwizdz\textsuperscript{1}, Dongji Feng\textsuperscript{2},\\
\textsuperscript{1} Databricks 
\\
\textsuperscript{2} MCS department, Gustavus Adolphus College\\
{\tt wenxie0010@gmail.com , gray.gwizdz@gmail.com, djfeng@gustavus.edu}}

\begin{document}
\maketitle
\footnotetext[1]{Work does not relate to position at Databricks}

\pagestyle{plain} 


\begin{abstract}


While Large Language Models (LLMs) have emerged as promising tools for evaluating Natural Language Generation (NLG) tasks, their effectiveness is limited by their inability to appropriately weigh the importance of different topics, often overemphasizing minor details while undervaluing critical information, leading to misleading assessments. Our work proposes an efficient prompt design mechanism to address this specific limitation and provide a case study. Through strategic prompt engineering that incorporates explicit importance weighting mechanisms, we enhance using LLM-as-a-Judge ability to prioritize relevant information effectively, as demonstrated by an average improvement of 6\% in the Human Alignment Rate (HAR) metric.

\end{abstract}

\section{Introduction}\label{sec:intro}


 Automated evaluation metrics like BERTScore~\cite{zhang2019bertscore} and BARTScore~\cite{yuan2021bartscore} offer scalable and cost-effective performance in evaluating Natural Language Generation (NLG) tasks. BERTScore, while leveraging pre-trained semantic embeddings, often struggles to capture contextual nuances and may overemphasize lexical similarity over factual accuracy or logical consistency~\cite{sellam2020bleurt,sai2022survey}. BARTScore, though designed to improve upon these limitations, still faces challenges in aligning with human judgments, particularly in tasks requiring domain-specific knowledge or complex reasoning.\cite{lu2022toward,liu2023llms}. As a result, while automated metrics provide efficiency and scalability, they often fall short of replicating the comprehensiveness and depth of expert-driven evaluations. This highlights the need for hybrid approaches that combine the strengths of human expertise with the scalability of automated methods.

As LLMs continue to evolve, employing them as judges, called ``LLM-as-a-Judge''  is becoming increasingly compelling. LLMs, trained on vast and diverse datasets, have developed a nuanced understanding of language, enabling them to assess the quality of generated content effectively. Their ability to provide consistent evaluations can mitigate the variability inherent in human assessments. Moreover, using LLMs for evaluation is scalable and efficient, allowing for rapid assessments across large datasets without the logistical constraints of coordinating human evaluators~\cite{dong-etal-2024-llm}.


Yet, LLMs rely on contextualized information and often prioritize conventional details from training data. However, in certain cases, these elements may not be crucial, leading the model to misjudge their relative importance 
\cite{petroni2019language,jiang2020can,lin2021truthfulqa}.
Also, LLMs struggle to assign appropriate weights to different types of errors, such as factual errors versus grammatical errors 
\cite{sai2022survey,chiang2023can,zheng2023judging}.



In the NLG system, for tech companies, the primary users are often professional software developers or domain experts. When these users interact with a search system, they typically have a general idea of the desired output. As they review the results, they inherently apply a weighting mechanism to prioritize the most relevant components and focus on the information that aligns with their specific needs. Currently, a growing trend among companies is to use LLM-as-a-Judge to evaluate the quality of generated text. These evaluations often involve comparing the model’s output to a gold standard and assessing whether it meets predefined criteria. While this approach has its merits, a significant limitation arises: LLMs frequently fail to accurately weigh the importance of different components in the text\cite{petroni2019language,jiang2020can,lin2021truthfulqa}. For instance, in a software development context, an LLM might overemphasize the importance of a software release date thus giving rejection, even when the user's main objective is to obtain details about a specific feature or functionality.
Such trivial misunderstandings significantly limit the usefulness of LLM-as-a-Judge in domain-specific contexts. 




In this paper, we present a case study that addresses the aforementioned issue through two key strategies: "Explicit Error Weighting" and "Prompt Engineering". In the first step, we follow the original setting to use LLM-as-a-Judge (without customized prompt). Then we calculate the Human Alignment Rate (HAR). 
In our second step judge, we tested different LLM-as-Judges individually with our customized prompt. On average, we achieve a 6.4\% improvement in HAR, demonstrating the effectiveness of this method in aligning LLM decisions with human judgments.



\section{Related Work}\label{sec:background}

\noindent\textbf{LLM-as-a-Judge}: Previous work has explored the use of LLM-as-Judges and established benchmarks for evaluating their performance. For instance, ~\cite{zheng2023judging} investigates how LLMs can assess the quality of chatbot responses and compares their performance to human judgments. Additionally, ~\cite{kojima2022large} demonstrates that LLMs can perform reasoning tasks without explicit training, a capability that is foundational for their use as judges in evaluating complex outputs. Specifically, ~\cite{beurer2023prompting} examines how prompt engineering can guide LLMs to perform specific tasks, including evaluation and judgment, by treating prompts as a form of programming.

\noindent\textbf{Distinct from previous work}: Our work introduces two key innovations that differentiate it from these prior studies: 1) While earlier research primarily relied on the inherent reasoning and evaluation capabilities of LLMs, our approach introduces a novel explicit error weighting mechanism. Unlike traditional methods that treat all errors or evaluation criteria uniformly, our system assigns dynamic weights to different types of errors based on their significance. This ensures that critical errors are prioritized, leading to more accurate and context-aware evaluations. 2) To conduct this mechanism, we encode a weighting algorithm into our prompts that explicitly guide the LLM to focus on the most important aspects of the task.

\section{Background}\label{sec:Background}
\vspace{-2mm}
\subsection{Minor Issue and Motivation}

Traditional evaluation metrics incorporate weighting algorithms directly into their formulas ~\cite{feng2023joint,lin2004rouge}, which limits scalability and generalizability. An alternative approach is prompt design, which allows for integrating algorithms as plain text or in different formats within the prompt itself. Our case study specifically examines the issue that LLMs exhibit a low tolerance for minor omissions—an aspect that, from a conventional perspective, is considered critical due to their training on vast corpora of conventional knowledge.





\subsection{Proposed Prompt Design}\label{sec:promptDesign}
To improve the performance of using "LLM-as-a-Judge", we propose the following framework. We consider the evaluating process is essentially assessing \textbf{``Factual Content''}, which has three types: 1) \textbf{``Critical Fact''}, 2) \textbf{``Supporting Fact''}, and 3) \textbf{``Trivial Facts''}. In this framework, a "Fact" is any verifiable piece of information within a response, while "Critical Facts" are the essential elements that directly address the user's query and whose absence significantly weakens the answer. "Supporting Facts" provide additional context and clarification that enrich the response without being strictly necessary for correctness, and "Trivial Facts" are minor details that, although they can enhance the narrative, are optional and their omission is not penalized. The evaluation prompt will utilize the Algorithm ~\ref{alg:ai_response_evaluation}. From this algorithm, we incorporated our defined three types of facts, taking into account semantic similarity, output format, and required explanation in the evaluation result. This prompt aligns with level 6 of the TELeR prompt design taxonomy ~\cite{santu2023teler}. 





\begin{algorithm}[!thb]\footnotesize
\caption{AI Response Evaluation Framework}
\label{alg:ai_response_evaluation}

\begin{algorithmic}
\Require Generated AI response (\texttt{AI\_Response}), Gold standard response (\texttt{Gold\_Response})
\Ensure Evaluation metrics:  
  \begin{itemize}
    \item \texttt{semantic\_similarity}
    \item \texttt{fact\_match\_ratio}
    \item \texttt{critical\_facts\_missed}
    \item \texttt{supporting\_facts\_missed}
    \item \texttt{trivial\_facts\_missed}
    \item \texttt{final\_score}
    \item \texttt{explanation}
  \end{itemize}

\State \textbf{Step 1: Compare Responses}  
    \State Compare the factual content of \texttt{AI\_Response} with \texttt{Gold\_Response}.

\State \textbf{Step 2: Fact Inclusion}  
    \State Identify critical and supporting facts in \texttt{Gold\_Response} and check if they exist in \texttt{AI\_Response}.

\State \textbf{Step 3: Additional Facts}  
    \State Allow \texttt{AI\_Response} to include additional relevant facts not present in \texttt{Gold\_Response}.

\State \textbf{Step 4: Tolerate Omissions}  
    \State Permit omissions of trivial facts in \texttt{AI\_Response} without penalty.

\State \textbf{Step 5: Generate Evaluation Metrics}  
    \State Compute evaluation metrics and format them into a JSON object.

\State \textbf{Step 6: Output Results}  
    \State Return the JSON object containing the computed evaluation metrics.

\end{algorithmic}
\end{algorithm}


\section{Experiment Design}\label{sec:Experiment}

\subsection{Dataset}

The Databricks Documentation Evaluation Set\footnote{\url{https://notebooks.databricks.com/demos/dbdemos-dataset/llm/databricks-documentation/databricks_doc_eval_set.parquet}} is a structured dataset designed to evaluate the effectiveness of the Retrieval-Augmented Generation (RAG) process in generating accurate and contextually relevant responses based on Databricks documentation. It consists of user queries (\texttt{request}), expected retrieved contextual information from official documentation (\texttt{expected\_retrieved\_context}) and corresponding high-quality, manually curated responses (\texttt{expected\_response}), which serve as the gold standard for evaluation. Each query is uniquely identified by a \texttt{request\_id}. Additionally, the responses (\texttt{response}) are generated using a Retrieval-Augmented Generation (RAG) system that integrates relevant contextual information, powered by the databricks-meta-llama-3-70b-instruct model.



\begin{table}[!thb]\footnotesize
\centering
\begin{tabular}{lcc}
\hline
\textbf{Statistic}               & \textbf{Request} & \textbf{Response} \\ \hline
Total Count                      & 192                         & 192                          \\
Average Length (chars)           & 124.03                      & 954.76                       \\
Min Length (chars)               & 53                          & 56                           \\
Max Length (chars)               & 1024                        & 5763                         \\
Average Word Count               & 21.04                       & 139.34                       \\ \hline
\end{tabular}
\caption{This table provides an overview of key statistics for requests and responses, including counts, character lengths, and average word counts.}
\label{tab:request_response_statistics}
\end{table}





\subsection{LLMs} 

For this experiment, we selected five of the most popular commercial large language models: OpenAI GPT-4o, GPT-4o-mini~\cite{achiam2023gpt}, Meta Llama3.1 two versions~\cite{touvron2023llama} and Mixtral-8x7B Instruct~\cite{jiang2024mixtral} . We utilized their respective APIs. Table~\ref{tab:LLMs} summarizes the details of these five LLMs. Our baseline model to perform the first step is a customized GPT-4o trained for our used dataset.

\begin{table}[!thb]\footnotesize
\centering
\begin{tabular}{|lllll|}
\hline
\multicolumn{5}{|l|}{LLM informations}                         \\ \hline
\multicolumn{2}{|l|}{LLMs used as Judges}  & \multicolumn{3}{l|}{ Model Size} \\ \hline

\multicolumn{2}{|l|}{Baseline (Customized GPT-4o)}  & \multicolumn{3}{l|}{Not Officially Disclosed}      \\ \hline

\multicolumn{2}{|l|}{OpenAI GPT-4o mini}  & \multicolumn{3}{l|}{Not Officially Disclosed}      \\ \hline
\multicolumn{2}{|l|}{OpenAI GPT-4o}  & \multicolumn{3}{l|}{Not Officially Disclosed}      \\ \hline
\multicolumn{2}{|l|}{Mixtral - 8x7B } & \multicolumn{3}{l|}{8 Experts, size of each expert is 7B}      \\ \hline
\multicolumn{2}{|l|}{MetaAI Llama3.1-databricks} & \multicolumn{3}{l|}{meta-llama-3-70b-instruct  (70B)}      \\ \hline
\multicolumn{2}{|l|}{MetaAI Llama3.1-databricks} & \multicolumn{3}{l|}{meta-llama-3-1-405b-instruct (405B)}      \\ \hline




\end{tabular}%
\caption{Large language models studied in this paper.}
\label{tab:LLMs}
\end{table}


\subsection{Evaluation Metric}

Human Alignment Rate (HAR) refers to how well an AI system, especially for LLM, aligns its outputs with human expectations, values, or judgments and is widely used in industry. It has also been used as a reward function for Reinforcement learning~\cite{kaufmann2023survey,kabir2025beyond} or ranking optimization~\cite{song2024preference}.
HAR equation~\ref{equ:HAR} can be seen below: 

\begin{equation}\label{equ:HAR}\small
\text{HAR} = \frac{\text{\# of AI outputs that match human judgments}}{\text{Total \# of evaluated outputs}} \times 100\%
\end{equation}

\section{Experimental Results}

\begin{figure*}[!hbt]\footnotesize
\centering
\includegraphics[width=0.95\textwidth]{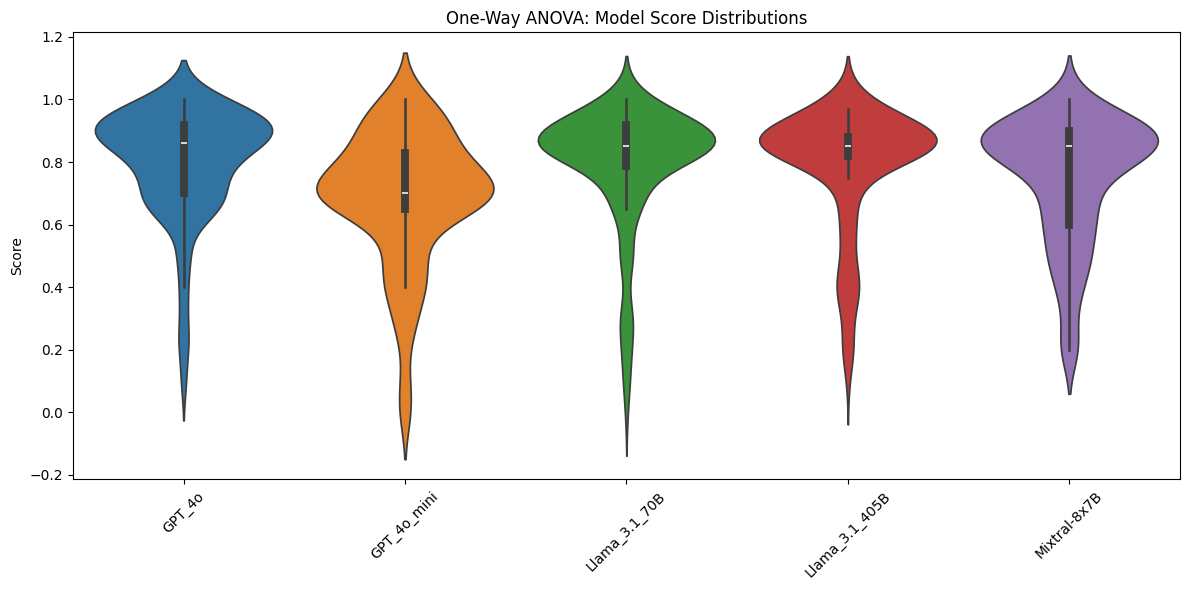}
\caption{One-Way ANOVA: Model Score Distributions.}
\label{fig:anova_violin}
\end{figure*}

\subsection{Human Alignment Rate}

 Table ~\ref{tab:model_results} compares the HAR (\%) of several models. The original baseline Model has an 85.9\% rate. In our two step mechanism, we require following LLMs to re-evaluate the generated results prompted by our designed prompt.
First, we observe GPT-4o achieves 94.3\%, and its smaller version, GPT-4o-mini, scores 88.5\%. We also tested two versions of LLama (customized version), Llama 3.1 70B Instruct reaches 89.1\%, and the larger Llama 3.1 405B Instruct improves to 93.8\%. Interestingly, the top performer is Mixtral-8x7B Instruct with 95.8\%, showing that our designed mechanism generally aligns better with human expectations.

\begin{table}[ht]\footnotesize
\centering

\centering
\begin{tabular}{|l|c|}
\hline
\textbf{\centering Model} & \textbf{Human Alignment} \\ 
                          & \textbf{Rate (\%)}       \\ \hline
Baseline Model          & 85.9                     \\ \hline
GPT-4o                    & 94.3                     \\ \hline
GPT-4o-mini               & 88.5                     \\ \hline
Llama 3.1 70B Instruct     & 89.1                     \\ \hline
Llama 3.1 405B Instruct    & 93.8                     \\ \hline
Mixtral-8x7B Instruct      & 95.8                     \\ \hline
\end{tabular}
\caption{Comparison of Model Results After Two Step LLM-as-a-Judge}
\label{tab:model_results}
\end{table}



\subsection{Model Score Distributions}
Figure~\ref{fig:anova_violin} compares the score distributions of tested five models. From this violin figure, we can observe that GPT-4o and Llama 3.1 405B show higher median scores and tight distributions. Their score profiles are steep and narrow around the median, indicating that these models not only perform well but also do so consistently with few outliers. In contrast, GPT-4o-mini and Mixtral-8x7B have lower median scores and broader distributions. The wider spread in their scores suggests greater variability, meaning these models may sometimes perform well but can also struggle with certain tasks, leading to occasional low scores.

The Llama 3.1 70B model appears to occupy an intermediate position, performing reasonably well with a slightly wider distribution compared to its 405B counterpart. This suggests that while it generally achieves competitive scores, it is more prone to fluctuations, reinforcing the observation that scaling up the model size contributes to both improved median performance and reduced variance.

The violin plot also captures subtle differences in the lower tail behavior of each model. GPT-4o-mini, for example, has a notably elongated lower tail, indicating that it is more susceptible to low-scoring outliers. This suggests that while it may occasionally achieve high scores, it also exhibits greater inconsistency. Similarly, Mixtral-8x7B, despite its relatively high median, demonstrates a greater spread, suggesting that it may be effective in some scenarios but is less robust overall.

Overall, this analysis highlights a clear trade-off between performance consistency and score variability. Models like GPT-4o and Llama 3.1 405B are preferable when reliability and stability are critical, as they consistently deliver high scores with minimal deviation. Conversely, GPT-4o-mini and Mixtral-8x7B, despite having competitive scores at times, introduce greater uncertainty, which could impact their suitability for applications where predictability and robustness are key considerations



\subsection{Statistical Significance Testing}
Table~\ref{tab:anova_tukey_results} provides One-way ANOVA and Tukey HSD Test Results as our statistical significance testing reveals notable differences in performance among HAR.

The One-way ANOVA test produced a significant F-statistic of 10.3042 and a p-value of 0.0000, indicating that at least one model's performance distribution differs significantly from the others. This confirms that the models do not perform identically and that there are observable differences in their scores.

From the pairwise comparisons shown in Table 4, we can observe that GPT-4o-mini demonstrates significant performance differences when compared to all other models. The adjusted p-values for comparisons involving GPT-4o-mini are consistently below the 0.05 threshold, indicating statistically significant differences. This suggests that GPT-4o-mini exhibits a distinctive performance profile that sets it apart from the other models, which may reflect underlying differences in its design or capabilities.

In contrast, GPT-4o does not show statistically significant differences when compared to Llama-3.1-405B, Llama-3.1-70B, or Mixtral-8x7B. The adjusted p-values for these comparisons are well above 0.05, suggesting that these models exhibit similar levels of performance. This similarity in performance among GPT-4o, Llama-3.1-405B, Llama-3.1-70B, and Mixtral-8x7B indicates that these models can be considered interchangeable for certain tasks or contexts where consistency across models is desired.

Further analysis of the Llama-3.1 variants and Mixtral-8x7B reveals that comparisons between these models yield non-significant differences. The adjusted p-values are consistently above the 0.05 threshold, indicating that these models perform comparably. This suggests that scaling from Llama-3.1-70B to Llama-3.1-405B does not lead to a statistically significant improvement in performance, despite the potential increase in computational resources or parameter count.

The contradiction between the violin plot and the statistical significance results highlights the distinction between observable performance trends and statistically validated differences. The violin plot suggests that scaling from Llama 3.1 70B to Llama 3.1 405B improves performance consistency. However, the Tukey HSD test results (p = 0.9984) show that this difference is not statistically significant. This discrepancy can arise because statistical tests primarily evaluate differences in mean performance rather than distributional stability, meaning a model could exhibit less variance without showing a statistically significant improvement in its average score. Despite the lack of statistical significance, the practical implications remain: Llama 3.1 405B demonstrates greater stability and lower variance, making it a more reliable choice for applications where consistent performance is crucial, even if its mean performance is not significantly different from that of Llama 3.1 70B. Whether the scaling is beneficial depends on the trade-off between prioritizing statistical significance versus practical model reliability, where the latter may still favor the larger model due to its reduced performance variability

These findings highlight important trade-offs when selecting models for different tasks. While GPT-4o-mini stands out with its unique performance characteristics, it also introduces a level of variability that may make it less predictable for certain applications. On the other hand, models such as GPT-4o, Llama-3.1-405B, and Mixtral-8x7B offer more consistent performance, making them suitable choices for scenarios where reliability and predictability are critical.

\begin{table}[!htb]\footnotesize
\vspace{0.5em}

\begin{tabular}{lcccccc}
\hline
\textbf{Group1}    & \textbf{Group2}       & \textbf{p-adj} \\ \hline
GPT-4o            & GPT-4o-mini                   & 0.0000                   \\
GPT-4o            & Llama-3.1-405B                 & 0.9392             \\
GPT-4o            & Llama-3.1-70B                & 0.9893                \\
GPT-4o            & Mixtral-8x7B                      & 0.0777                  \\
GPT-4o-mini      & Llama-3.1-405B                 & 0.0000              \\
GPT-4o-mini      & Llama-3.1-70B                    & 0.0000               \\
GPT-4o-mini      & Mixtral-8x7B                       & 0.0254            \\
Llama-3.1-405B   & Llama-3.1-70B                & 0.9984                \\
Llama-3.1-405B   & Mixtral-8x7B                     & 0.3762           \\
Llama-3.1-70B    & Mixtral-8x7B                 & 0.2263              \\ \hline
\end{tabular}%
\caption{Pair Wise Statistical Significance Difference Testing}
\label{tab:pair_wise}
\end{table}

\vspace{-3mm}

\begin{table}[htb]\footnotesize
\centering
\begin{tabular}{lcc}
\hline
\textbf{Statistic} & \textbf{Value} \\ \hline
F-statistic        & 10.3042        \\
p-value            & 0.0000         \\ \hline
\end{tabular}%
\caption{One-way ANOVA and Tukey HSD Test Results}
\label{tab:anova_tukey_results}

\end{table}

\section{Conclusion }\label{sec:conclusion}

In this work, we introduce a two-step mechanism designed to enhance using LLM-as-a-Judge. Our approach incorporates a weighting algorithm into the prompt, which guides the LLM during the second-step evaluation process.  We conduct a case study using a dataset specialized in software engineering and evaluate its performance across five widely used LLMs. Our proposed method achieves an average improvement of 6\% in HAR. Particularly, Mixtral-8x7B Instruct emerged as the clear winner in this competition, achieving a 95.8\% HAR, outperforming all other LLMs, including two customized Llama models.


\section{Limitations}

Our case study demonstrates the effectiveness of the proposed method on a single dataset, and the prompts used in this study were manually designed to address the unweighted evaluation issue. While the results are promising, the approach may face scalability challenges until it is tested on a broader range of datasets. Additionally, future work should explore the use of auto-generated prompts to improve efficiency and reduce reliance on manual design. AI assistant was utilized in the writing process.


\bibliography{custom}
\bibliographystyle{acl_natbib}

\newpage
\newpage
\section{Appendix}

\subsection{Example }

We provide one example in our used dataset.
\textbf{Request:} What is the syntax for using the schema\_of\_csv function in Databricks SQL?

\textbf{Response:} The syntax for using the schema\_of\_csv function in Databricks SQL is:
schema\_of\_csv(csv [, options]). Where CSV is a STRING literal with valid CSV data, and options is an optional MAP literal where keys and values are STRING. The function returns a STRING composing a struct, with field names derived by position as \_Cn and values holding the derived formatted SQL types. For details on options, see the from\_csv function.

\textbf{Expected Response:} The syntax for using the schema\_of\_csv function in Databricks SQL is:
schema\_of\_csv(csv [, options]).
Where csv is a STRING literal with valid CSV data, and options is an optional MAP literal where keys and values are STRING. The function returns a STRING composing a struct. The field names are derived by position as \_Cn. The values hold the derived formatted SQL types. For details on options, see the from\_csv function.

\subsection{Exact Prompt }

\begin{quote}
\textit{
You are an AI judge evaluating the quality of an AI-generated response compared to a gold standard response. Your task is to determine if the AI response matches the gold response based on the following criteria:
}
\begin{enumerate}
    \item Compare the factual content of both responses.
    \item Check if the AI response includes facts that are also present in the gold response.
    \item The AI response can have additional facts not present in the gold response.
    \item The AI response should not miss any critical or supporting facts from the gold response.
    \item The AI response can miss trivial facts from the gold response.
\end{enumerate}
\textit{
Please analyze the following responses:
}
AI Response: \{ai\_response\}

Gold Response: \{gold\_response\}

\textit{
Provide your evaluation in the following JSON format:
}

\begin{tabular}{@{}ll@{}}
\{ & \\
    & "semantic\_similarity": \textless float between 0 and 1\textgreater, \\
    & "fact\_match\_ratio": \textless float between 0 and 1\textgreater, \\
    & "critical\_facts\_missed": \textless integer\textgreater, \\
    & "supporting\_facts\_missed": \textless integer\textgreater, \\
    & "trivial\_facts\_missed": \textless integer\textgreater, \\
    & "final\_score": \textless float between 0 and 1\textgreater, \\
    & "explanation": \textless string explaining your evaluation\textgreater \\
\} & \\
\end{tabular}
\end{quote}

\end{document}